 \documentclass[pmlr,twocolumn]{jmlr} % W&CP article

\usepackage{booktabs}
\usepackage[load-configurations=version-1]{siunitx} % newer version
\usepackage{graphicx}

\theorembodyfont{\upshape}
\theoremheaderfont{\scshape}
\theorempostheader{:}
\theoremsep{\newline}

\jmlrvolume{LEAVE UNSET}
\jmlryear{2020}
\jmlrsubmitted{LEAVE UNSET}
\jmlrpublished{LEAVE UNSET}
\jmlrworkshop{Machine Learning for Health (ML4H) 2020} % W&CP title

\title[Addressing the Real-world Class Imbalance Problem in Dermatology]{Addressing the Real-world Class Imbalance Problem in Dermatology}

\author{%
\Name{Wei-Hung Weng}\thanks{Work done at Google.} \Email{ckbjimmy@mit.edu}\\
\addr MIT
\AND
\Name{Jonathan Deaton} \Email{jdeaton@google.com} \\
\addr Google Health
\AND
\Name{Vivek Natarajan} \Email{natviv@google.com} \\
\addr Google Health
\AND
\Name{Gamaleldin F. Elsayed} \Email{gamaleldin@google.com} \\
\addr Google Research
\AND
\Name{Yuan Liu} \Email{yuanliu@google.com} \\
\addr Google Health
}

\begin{document}

\raggedbottom

\maketitle

\begin{abstract}
  Class imbalance is a common problem in medical diagnosis, causing a standard classifier to be biased towards the common classes and perform poorly on the rare classes. This is especially true for dermatology, a specialty with thousands of skin conditions but many of which have low prevalence in the real world. Motivated by recent advances, we explore few-shot learning methods as well as conventional class imbalance techniques for the skin condition recognition problem and propose an evaluation setup to fairly assess the real-world utility of such approaches. We find the performance of few-show learning methods does not reach that of conventional class imbalance techniques, but combining the two approaches using a novel ensemble improves model performance, especially for rare classes. We conclude that ensembling can be useful to address the class imbalance problem, yet progress can further be accelerated by real-world evaluation setups for benchmarking new methods.
\end{abstract}

\section{Introduction}
\begin{figure}[htbp]
\centering
\includegraphics[width=1\linewidth]{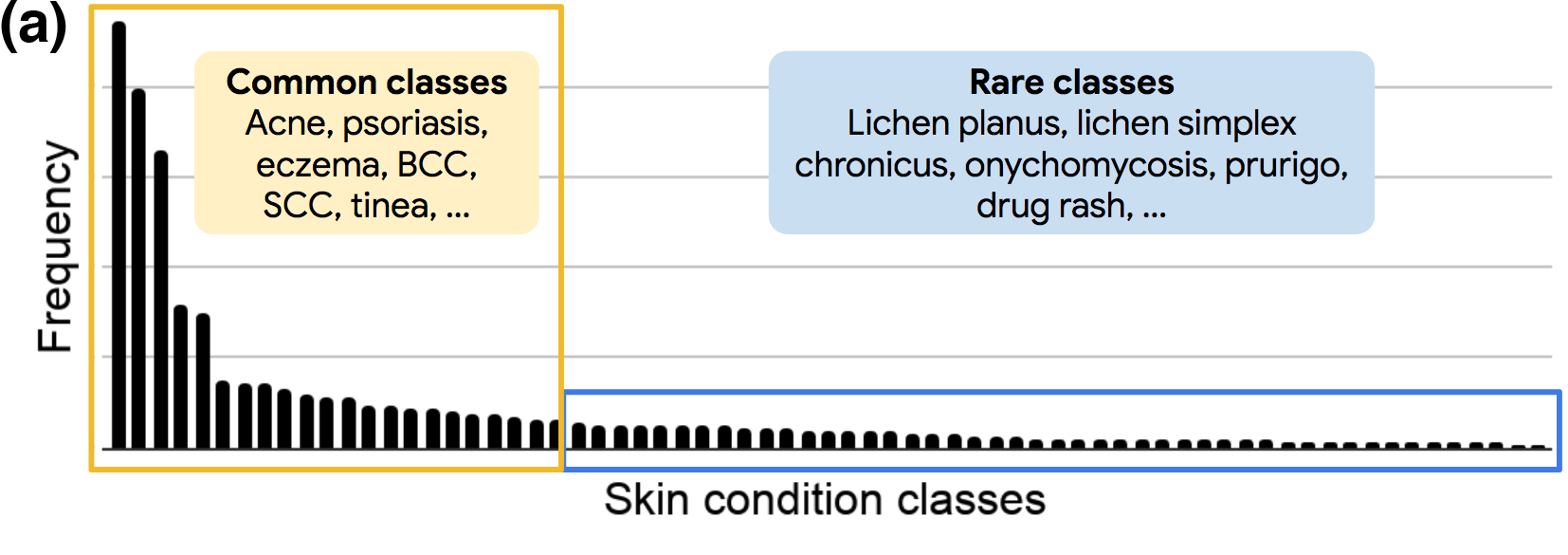} 
\vspace{5pt}
\includegraphics[width=1\linewidth]{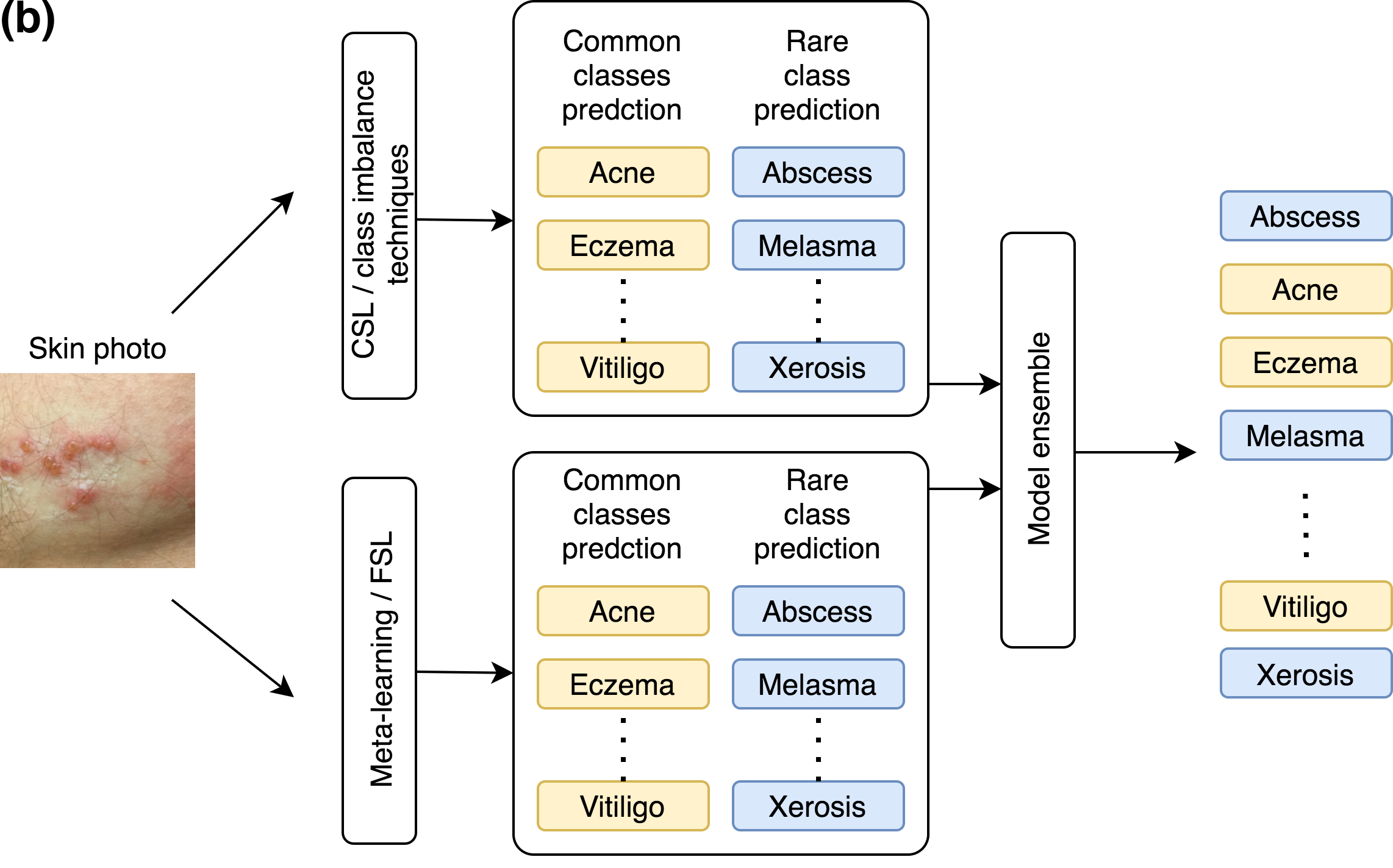} 
\hfill
\vspace{-25pt}
\caption{Class imbalance problem in dermatology at a glance (a) and proposed modeling framework (b).}
\label{fig1}
\vspace{-20pt}
\end{figure}

Skin disease is the fourth leading cause of non-fatal medical conditions burden worldwide \citep{seth2017global}. 
Due to the global shortage of dermatologists, access to dermatology care is limited, leading to rising costs, poor patient outcomes, and health inequalities.
Recent research endeavors have demonstrated that deep learning systems built to detect skin conditions from either dermatoscopic or digital camera images can achieve expert level performance in diagnosing certain skin conditions \citep{esteva2017dermatologist, liu2020deep}. 
Despite the encouraging progress, such systems can only identify a few common skin conditions well, leaving a vast number of skin conditions still unaddressed in the real-world (Figure~\ref{fig1} (a)). 

As many skin conditions occur infrequently in real-world, datasets collected from a natural patient population are highly imbalanced, with few examples for those skin conditions. 
This makes it challenging to build models that can reliably detect rare skin conditions. 
Such limitations not only may diminish the clinical utility of these systems due to the low skin condition coverage, but also may cause harm to patients as they are often biased towards common diseases.

A wide variety of techniques have been proposed over the years to address the class imbalance problem in machine learning. 
These include relatively classic approaches like modified resampling strategies \citep{chawla2002smote}, loss reweighting \citep{eban2017scalable}, focal loss and bias initialization \citep{lin2017focal}, to more complex approaches such as using generative models to augment the rare classes with synthetic images \citep{ghorbani2020dermgan}.

A related counterpart to the class imbalance problem is \emph{few-shot learning} (FSL), a learning scenario inspired by the human ability to learn from very few examples~\citep{lake2011one}.
It can further be generalized to the meta-learning framework that aims to gain learning experiences from solving many meta tasks in order to achieve a better performance for the tasks (in this case, rare classes) with limited training examples~\citep{vinyals2016matching}.

The classic evaluation framework for FSL follows an ``$N$-way-$k$-shot'' setting: given $k$ training examples for $N$ classes as a random subset of all classes, test if the FSL techniques can correctly distinguish among the $N$ classes. 
It is often not understood whether FSL will translate well beyond this contrived setting to the \emph{real-world} problem, when the task requires recognizing the correct class among \emph{all} possible classes (all-way classification) in the wild~\citep{chen2019closer,triantafillou2019meta}. 
Furthermore, it is challenging to compare the classification performance between FSL and the conventional supervised training without a unified evaluation framework.

In this paper, we investigate various FSL methods to tackle the class imbalance problem in skin condition classification (Figure~\ref{fig1} (b)). 
We modify the typical FSL evaluation setup to adapt to the real-world setting, and design studies to compare FSL methods to \emph{conventional supervised learning} (CSL) with various classic techniques to address the class imbalance problem. We further explore the potential use of the combination of FSL and CSL with various class imbalance techniques, motivated by the hypothesis that a combination may provide independent predictions, capturing different aspects of the data. 
Specifically, our contributions are:
\begin{itemize}
    \item We propose a real-world evaluation framework to compare the FSL methods against conventional methods (i.e., CSL with class imbalance techniques) for the class imbalance problem.
    \item We find that FSL methods don't outperform CSL with class imbalance techniques, yet an ensemble of these two types can outperform either alone, especially for the rare classes.
\end{itemize}

\vspace{-10pt}
\section{Related Work}

\subsection{Class Imbalance}
\begin{figure*}[htbp]
\centering
\includegraphics[width=1\linewidth]{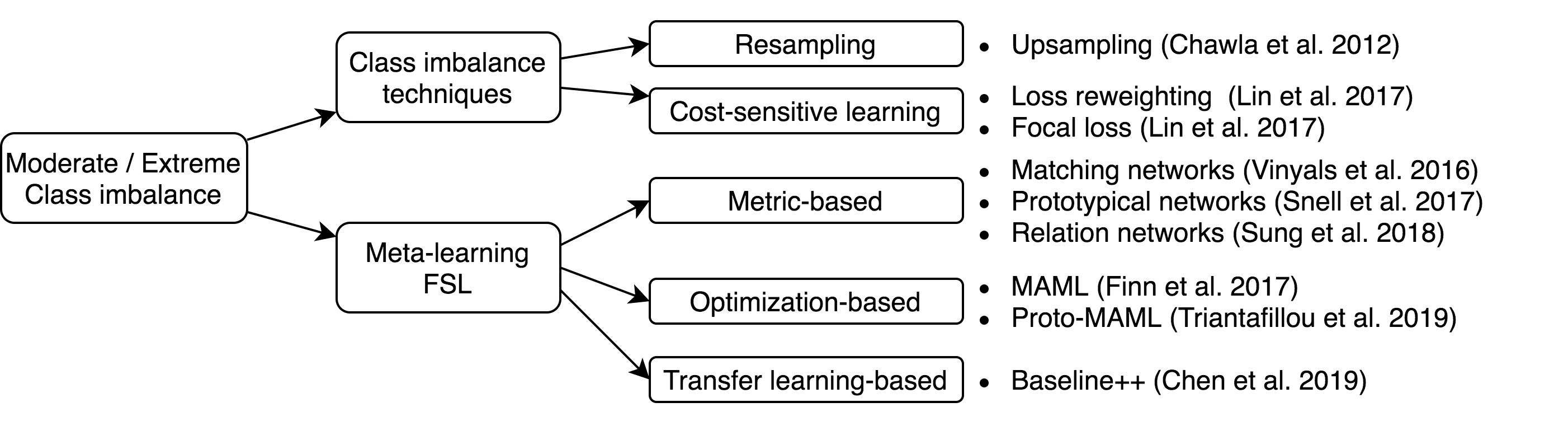} 
\hfill
\vspace{-25pt}
\caption{Categories of methods for tackling the class imbalance problem.}
\label{fig2}
\vspace{-15pt}
\end{figure*}

Imbalanced class distribution is common in real-world datasets. 
For mild class imbalance, machine learning algorithms, such as support vector machines~\citep{cortes1995support}, random forests~\citep{breiman2001random}, and modern deep learning techniques can usually handle such cases well. 
However, special care is required to manage moderate or extreme imbalance situations (Figure~\ref{fig2}), when minority classes constitutes typically less than 20\% or 1\% in the training data respectively.

\paragraph{Conventional Techniques}
Researchers have developed various resampling and cost-sensitive learning strategies to improve models under such skewed class distribution settings. 
Resampling aims to balance the class distribution in the training data by either downsampling common classes or upsampling rare classes. Downsampling may increase variance; thus, upsampling is more commonly used~\citep{chawla2002smote}, yet upsampling procedures for extremely imbalanced data may be computationally expensive~\citep{weng2019multimodal}.

In contrast, cost-sensitive learning penalizes algorithms by increasing the cost of classification mistakes on the rare classes.
It can be implemented in various ways, such as reweighting the losses of specific classes, introducing bias as prior into the classification loss at each class~\citep{lin2017focal}, or less used approaches such as global objectives~\citep{eban2017scalable}.
To prevent the easy examples in the common classes from dominating the gradients during classifier training, the focal loss function is developed~\citep{lin2017focal}, to not only out-weight the rare classes, but also emphasize the hard examples during training.

\paragraph{Few-shot Learning (FSL)}
FSL can also be utilized to improve the classification performance for the classes with very few training examples. 
The FSL algorithms can be categorized into several types: metric-based, optimization-based, and transfer learning-based approaches.

Metric-based FSL learn a representation by learning to compare training examples.
\citet{koch2015siamese} developed the Siamese neural network to compare two examples at the same time with two identical twin networks sharing the same weights. 
Matching network utilizes the cosine similarity as the distance metric and adopts the long-short term memory (LSTM) network for generating embeddings~\citep{vinyals2016matching}. 
Prototypical network, on the other hand, uses the Euclidean distance and convolutional networks to determine the embedding of the class prototypes~\citep{snell2017prototypical}. 
Relation networks further employ the concatenation strategy and relation module to score example similarity~\citep{sung2018learning}.

Optimization-based FSL aims to learn a set of parameters that allows a meta-learner to quickly adapt to new tasks~\citep{finn2017model}. 
\citet{ravi2016optimization} designed an LSTM-based meta-learner to replace the stochastic gradient decent algorithm in order to learn a better optimizer. 
One can also learn a better model initialization for faster task adaptation with less gradient updates (Model-Agnostic Meta-Learning, MAML)~\citep{finn2017model}. 
Others have integrated both metric and optimization-based FSL to further improve the performance~\citep{triantafillou2019meta}.

Finally, transfer learning tackles the few-shot classification fine-tuning a model pretrained on a much larger dataset~\citep{chen2019closer,tian2020rethinking}. 
\citet{chen2019closer} adopts the training-then-finetuning process for few-shot classification without meta-learning (Baseline++). 
Recently,~\citet{tian2020rethinking} demonstrated that transfer learning under the meta-learning framework can yield strong performance for the few-shot classification problem.

Despite the progress in FSL, they are typically evaluated in a contrived setting and little is known about how they work in real-world all-way classification problems. 
In this study, we therefore propose an evaluation method that allows us to use FSL for the all-way classification problem, and compare with the CSL-based methods.

\vspace{-10pt}
\subsection{Machine Learning and FSL in Dermatology}
Artificial intelligence in dermatology is a rapidly growing topic of interest in recent years~\citep{cruz2013deep,esteva2017dermatologist,liu2020deep,yang2018clinical,prabhu2019few,li2019difficulty,mahajan2020meta,guo2020broader,le2020transfer}.
For example,~\citet{esteva2017dermatologist} applied deep learning to clinical skin photos for two binary skin cancer classification tasks.
\citet{liu2020deep} developed a CSL-based system that identifies 26 common skin conditions with performance superior to general practitioners and on par with dermatologists.

However, it is challenging to extend such systems to support the rare skin conditions due to the limited training examples.
Previous efforts have explored various approaches, such as adopting domain knowledge to learn a better representation~\citep{yang2018clinical}, developing or modifying the meta-learning based methods~\citep{prabhu2019few,li2019difficulty,mahajan2020meta}, and making comparison between different approaches tackling the extreme class imbalance problem in dermatology~\citep{guo2020broader,le2020transfer}.
Yet there is no study investigating both the FSL and CSL-based class imbalance techniques and comparing them under the real-world all-way classification setting.

Our work focuses on the all-way classification under a extreme long-tailed data distribution that often occurs in real-world medical imaging tasks. Under an unified evaluation framework, we make comparison between FSL and CSL-based class imbalance techniques.
We further propose ensembling FSL with conventional methods and demonstrate gains from combining both strategies.

\vspace{-10pt}
\section{Methods}
In this work, we investigate the model performance on skin condition classification problem across CSL baseline, CSL with class imbalance techniques, FSL techniques, and different ensembles between these approaches. 
We use an evaluation setup that allows us to compare across all methods for the real-world classification problem, which entails properly addressing all skin conditions in our datasets.
In this section, we explain how we modify the evaluation setup to adapt FSL to this real-world scenario, and introduce the learning algorithms and metrics used for evaluation.

\vspace{-10pt}
\subsection{FSL Task and Evaluation Setup}

\begin{figure*}[htbp]
\centering
\includegraphics[width=1.0\linewidth]{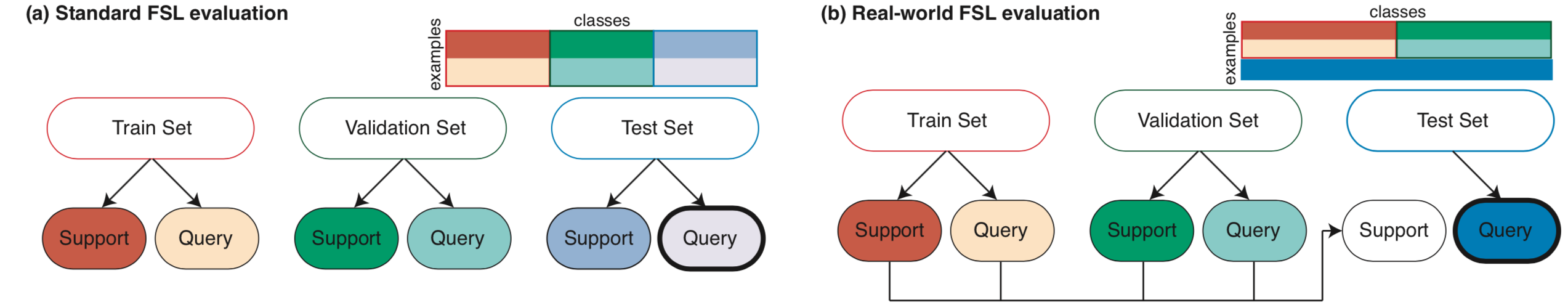} 
\hfill
\vspace{-20pt}
\caption{Differences between the standard (a) and the real-world (b) FSL evaluation settings. We report metrics over test query sets illustrated with bold box.}
\label{fig3}
\vspace{-20pt}
\end{figure*}

\paragraph{Task Formulation and Standard Evaluation}
FSL follows the meta-learning setting, which consists of training, validation, and testing phases: training for learning a classifier (meta-learner), validation for hyperparameter tuning, and testing for evaluating the learned classifier.
Data used in each of the three phases needs to be split into support and query set.
For the standard FSL evaluation, the classes used in these three phases should be all disjoint (Figure~\ref{fig3} (a)).

Given a dataset with $M$ classes, the training can be batch or episodic.
For batch training, we train a model using all examples from a random $N$ classes ($N < M$) in the train set; for episodic training, we train a classifier under $N$-way-$k$-shot learning scenario: 
in each training episode, we choose random $k$ samples from the random $N$ classes ($N < M$) in the train set to form a support set to learn a model, and then evaluate on a query set which includes other samples in the same $N$ classes in the training set. 
The validation and test sets are used to evaluate the model during and after training, respectively.

\paragraph{Real-world Evaluation for FSL}
The standard $N$-way-$k$-shot framework doesn't fit the real-world classification problem, where we need to discriminate all classes simultaneously.
Therefore, we introduce the following changes to the classic FSL evaluation setup (Figure~\ref{fig3} (b)):

(1) The dataset is split into development (train $\cup$ validation) and test sets in advance, where both sets may contain samples from all possible classes. 

(2) Different from the classical setup where the classes in all three splits are disjoint, in the real-world setup, only training and validation have disjoint classes, and both the support set and the query set for those classes come from the development set. 
During testing, the support set that is used to train the final model comes from the development set, whereas the query set is exactly the same as the test set and includes disjoint samples from \emph{all classes}. 

\vspace{-10pt}
\subsection{Modeling}
The following FSL algorithms are used in the study (Figure~\ref{fig2}): k-nearest neighbors (k-NN) and Baseline++~\citep{chen2019closer}, matching networks~\citep{vinyals2016matching}, prototypical networks~\citep{snell2017prototypical}, relation networks~\citep{sung2018learning}, MAML~\citep{finn2017model} and Proto-MAML~\citep{triantafillou2019meta}.
We also implemented the following CSL-based class imbalance techniques: upsampling with uniform sampling based on the ground truth class during training, bias initialization (BI) with the log of frequency in the training set~\citep{lin2017focal}, inverse frequency weighting (IFW), focal loss (FL)~\citep{lin2017focal}, and the combination of BI/IFW and FL/IFW (Appendix A for details).

\vspace{-10pt}
\subsection{Metrics}
We report the balanced accuracy (a.k.a. normalized sensitivity, or macro recall) separately for the common, rare, and all classes for the real-world evaluation. 
The balanced accuracy is used to account for the class imbalance issue to avoid overweighting the common classes.
We also reported top-1 all-way accuracy in the Appendix.
We further report the 95\% binomial confidence interval, derived from the mean $\mu$ and standard deviation $\sigma$ of accuracies of $E$ episodes of FSL or $E$ runs of the CSL-based model, which can be expressed as $\mu \pm 1.96 \frac{\sigma}{\sqrt{E}}$.

\vspace{-10pt}
\section{Experiments}

% \vspace{-10pt}
\subsection{Data}
We use the clinical skin images dataset collected by a teledermatology service serving 17 different clinical sites~\citep{liu2020deep}. 
We divide the data to different subsets as illustrated in Figure \ref{fig3} according to the temporal order (75\% in the development set and 25\% in the test set). Specifically, the test set is chosen to the more recent patient visits, to mimic the real world setting as using earlier data for model training and deploying the model to serve future patients.
The development set is further partitioned into train and validation sets based on per skin condition stratified sampling to ensure enough samples for training and validation.
The statistics of the data splits are shown in Table~\ref{tab-a1}. 

There are a total of 419 skin conditions in the dataset.
We define the common classes based on the selection criteria used in~\citep{liu2020deep} (more than 100 cases in the development set), and the rare classes as those with (1) more than 20 cases in the development set, and (2) more than 5 cases in the test set.
The criteria was established in order to ensure sufficient examples for training and evaluation. 
For other extremely rare classes (i.e., classes with $<$20 samples in the development set or $<$5 samples in the test set) (351 classes), we group them into a single aggregated category ``Other'' that belongs to a common class due to its sample size. 
In summary, we have 27 common classes, 42 rare classes (Figure~\ref{fig1} (a), Table~\ref{tab-a2}).

\vspace{-10pt}
\subsection{Training Frameworks}
The input for any model is an image of size of $448\times448$ , and the output is the corresponding skin condition prediction.
The Inception-V4 backbone is used for the CSL baseline, and CSL with class imbalance techniques.
Inception-V4 is evaluated as the most performant architecture in an internal neural architecture search experiment.

For FSL, we adopt the meta-dataset with the ResNet-18 backbone, which is one of the best performant network architectures in various FSL studies~\citep{triantafillou2019meta,tian2020rethinking}.
We train each model using Adam optimizer for 75000 steps with the exponential decay.
We follow~\citet{triantafillou2019meta} for the learning rate and weight decay setups.
We used the batch size of 64 and standard operations like random brightness, saturation, hue, contrast normalization, flip, rotation, and bounding box cropping for data augmentation.

\vspace{-10pt}
\section{Results}
We performed experiments to understand (1) whether FSL can be viably applied to the skin condition classification task, (2) how FSL compares against current CSL-based class imbalance techniques, especially on the rare classes, (3) whether combining CSL-based techniques and FSL can further improve the performance across the different skin condition classes.

\vspace{-10pt}
\subsection{Standard FSL Evaluation}
\begin{figure}[htbp]
\centering
\includegraphics[width=1\linewidth]{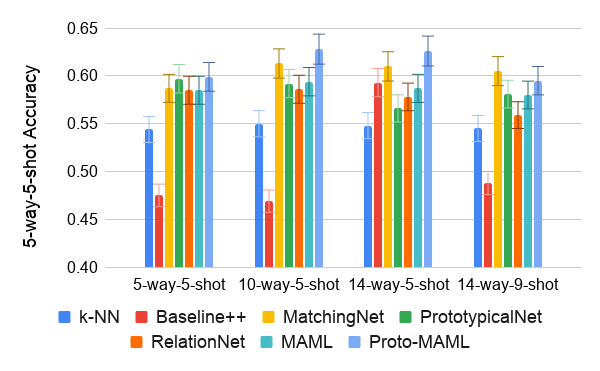} 
\hfill
\vspace{-30pt}
\caption{Standard FSL evaluation (5-way-5-shot accuracy with 95\% confidence interval) on the teledermatology dataset. We investigate the change of $N$ and $k$ values during training. x-axis is the FSL setting during training.}
\label{fig4}
\vspace{-25pt}
\end{figure}

In Figure~\ref{fig4} and Table~\ref{tab-a3}, we first benchmark and examine the feasibility of adopting the standard FSL evaluation (5-way-5-shot classification) to the teledermatology dataset.

We find that increasing $N$ yields a trend of better performance. 
For this particular dermatology use case, a higher way is a relatively optimal option to achieve the best performance under the standard FSL evaluation (Figure \ref{fig4}; $N$-way-5-shot bars where $N=5, 10, 14$). 
However, increasing $k$ is not helpful, possibly because of overfitting (Figure \ref{fig4}; 14-way-$k$-shot bars where $k=5,9$). 
This finding is consistent with the previous literature~\citep{snell2017prototypical}.
Among all methods, Proto-MAML and prototypical networks are comparable and consistently outperforms the others for the teledermatology dataset.
This dataset has shown similar properties as other FSL datasets under the standard FSL evaluation.

\vspace{-10pt}
\subsection{Real-world Evaluation}

\begin{figure*}[htbp]
\centering
\includegraphics[width=1\linewidth]{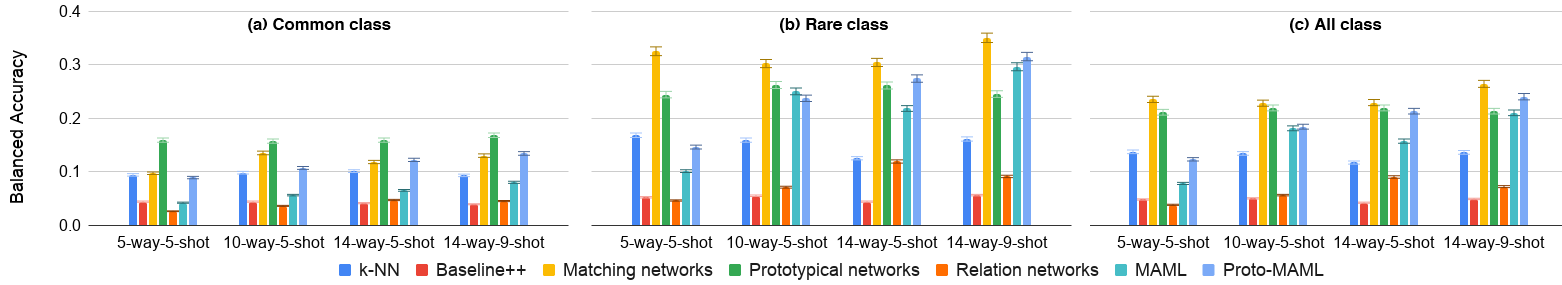} 
\hfill
\vspace{-25pt}
\caption{Real-world FSL evaluation. We report the balanced accuracy for (a) common class (b) rare class (c) all class prediction using different $N$-way-$k$-shot settings.}
\label{fig5}
\vspace{-30pt}
\end{figure*}

\paragraph{FSL-based Approaches}
In the real-world setting, we report all-way (69-way) performance (Figure~\ref{fig5}, Table~\ref{tab-a4}) on the test set (i.e. query set of the testing phase; see Figure \ref{fig3} (b)). 

The balanced accuracy is chosen as the metric to avoid overweighting the common classes.
We conduct the $N$-way-$k$-shot experiments again for finding the optimal training setting under the real-world setup.
Different from the findings in the standard FSL evaluation, we find that the optimal $N$ and $k$ values are not quite consistent across methods. 
The best performant method in the real-world FSL evaluation is also matching networks.
In Figure~\ref{fig5} (a, b) and Table~\ref{tab-a4}, we also find that the FSL methods consistently perform better on the rare classes than the common classes.

In brief, we find that the conclusion from the standard FSL evaluation that higher $N$-way training yields better meta-learner is inconsistent with the real-world evaluation results. 
Thus, more in-depth exploration is required in real world to identify the optimal $N$-way setting.
We also confirm our hypothesis that the FSL models can be beneficial for rare class prediction but may not be for common classes. 
For the common class prediction, relying on the CSL methods may be the better option in this case.
We later use the top performant models based on validation, which is the matching network model trained in 14-way-9-shot setting, for the model ensemble experiments below.

\begin{figure}[htbp]
\centering
\includegraphics[width=1.0\linewidth]{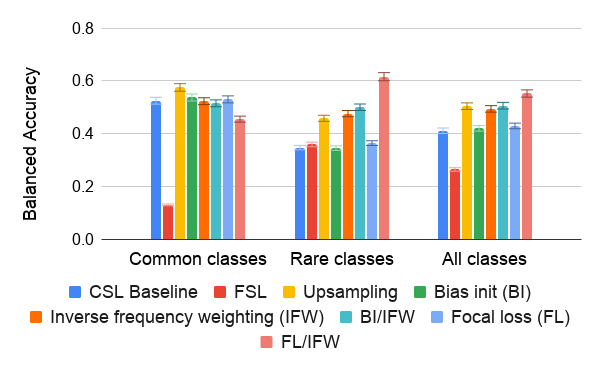}
\hfill
\vspace{-25pt}
\caption{Comparison between FSL and CSL-based class imbalance techniques on the all-way skin condition classification.}
\label{fig6}
\vspace{-25pt}
\end{figure}

\paragraph{CSL-based Class Imbalance Techniques}
Next, we compare the real-world all-way classification performance between FSL methods and CSL-based class imbalance techniques under the single model (non-ensemble) setting.
In Figure~\ref{fig6} and Table~\ref{tab-a5}\footnote{We further included results for prior correction, another class imbalance approach, in Appendix Table A5, and the method description in Appendix A.1.}, we find that the best FSL model (matching network) is only slightly better than the CSL baseline for the rare class prediction, possibly because of less training data. 
Yet the CSL-based class imbalance techniques yield better balance between common and rare classes, as demonstrated with all classes balanced accuracy.
Moreover, the CSL-based model integrating focal loss and inverse frequency weighting (FL/IFW) yields even better performance on the rare class classification. 
Such results suggests that FSL is not beneficial when used independently in this real-world setting.
One possible explanation could be that given FSL models are not using all the data from common classes during training, it hinders the learning of good low-level visual features, whereas CSL-based models are able to take advantage of it.

\paragraph{Model Ensemble}
To better utilize the FSL and CSL-based class imbalance techniques, we experiment on ensembling the trained models.
We approximate the joint ensemble output probability by computing the geometric mean of the probabilities across $M$ selected models: 
$$\textsc{P}_c = \left(\prod_{m=1}^{M} \textsc{P}_c^{(m)}\right)^{\frac{1}{m}}$$ 
where $\textsc{P}_c^{(m)}$ is the probability the $m^{th}$ model in the ensemble gives to class $c$. We compute the corresponding ensemble logits by first normalizing each model logits ($f_c^{(m)}$) using $\text{LogSumExp}(\cdot)$ (note this normalization does not alter the probabilities) before aggregating the logits from different models: 
$$\bar{f_c} = \frac{1}{M} \sum_{m=1}^{M} \left(f_c^{(m)} - \log \left(\sum_{i=1}^C \exp \left(f_i^{(m)}\right)\right)\right)$$
where $\bar{f_c}$ is the normalized logit for the ensemble.
We apply $\text{Softmax}(\cdot)$ to the normalized logits to compute the final ensemble probabilities.

\begin{figure*}[htbp]
\centering
\includegraphics[width=1\linewidth]{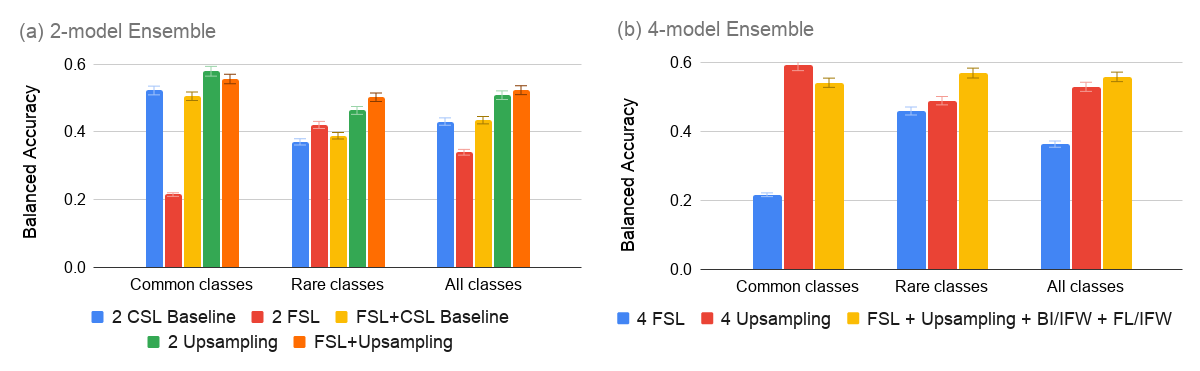}
\hfill
\vspace{-30pt}
\caption{Model ensemble. (a) 2-model ensemble to show the improvement after using FSL. (b) 4-model ensemble to demonstrate the improvement after ensembling models with different mechanisms.}
\label{fig7}
\vspace{-25pt}
\end{figure*}

To utilize the CSL-based methods in common class prediction, and FSL in rare class prediction, we further use the prediction from CSL-based methods if the ensemble prediction falls into the group of common classes, and use the prediction from FSL if it is from the group of rare classes while ensembling FSL with CSL-based models.
Based on our hypothesis and the results on the validation set, we use the best performant FSL model for rare classes, which is matching network, and the best CSL-based class imbalance model for common classes, which is upsampling method, as basic components for ensemble.
In Figure~\ref{fig7} (a) and Table~\ref{tab-a6}, we find that FSL-only ensembles do not perform well, which is consistent with the observations from the previous single model setting.
In contrast, ensembling FSL with CSL-based methods leads to a slight decrease in the common classes yet some improvement over rare classes and all classes in terms of balanced accuracy (Figure~\ref{fig7});  
ensembling the matching network model with the CSL model with upsampling technique (Figure~\ref{fig7} (a) orange) tends to strike a better balance between common and rare classes.
In Figure~\ref{fig7} (b), we find that making the ensemble more heterogeneous by using models with different mechanisms yields even better trade-off between common and rare classes along with a more notable performance increase, especially for rare classes.
These findings confirm our hypothesis that the ensemble of FSL and CSL can benefit such real-world imbalanced skin condition classification problem.

\vspace{-10pt}
\subsection{Qualitative Analysis}
\begin{figure*}[ht]
\centering
\includegraphics[width=1.0\linewidth]{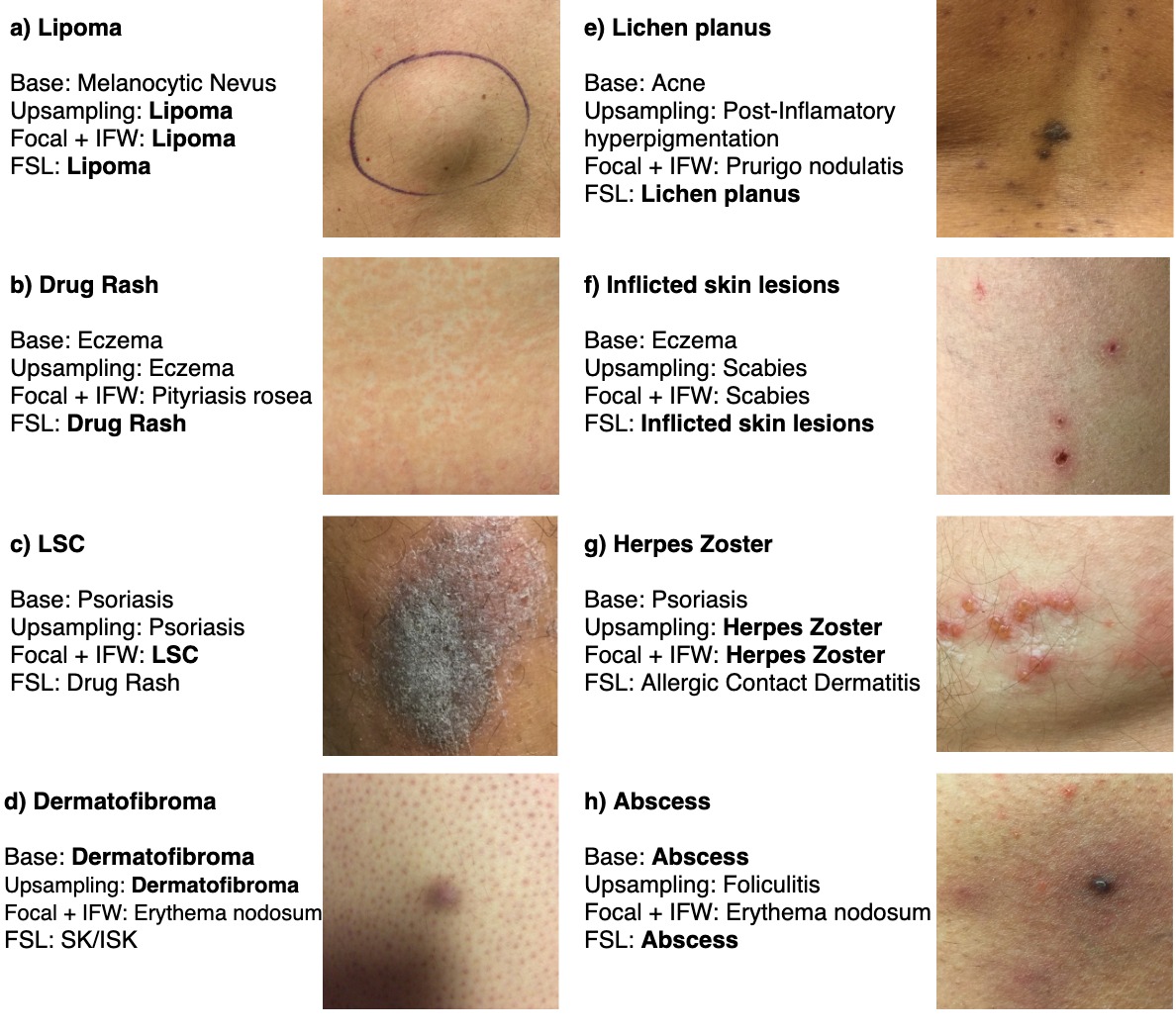}
\hfill
\vspace{-15pt}
\caption{Case study of rare classes. FSL and class imbalance techniques are better at identifying rare classes compared to the baseline model.}
\vspace{-25pt}
\label{fig8}
\end{figure*}

In Figure~\ref{fig8}, we present eight examples from rare classes alongside with predictions of different models.
We show that the FSL and CSL-based class imbalance techniques predict the rare classes more accurately while the CSL baseline tends to be biased toward the common classes such as acne, eczema and psoriasis.
For example, eczema is a very common skin condition with various presentations, but it can be visually similar to drug rash or other skin lesions. While the baseline tends to over-predict common classes, the FSL can correctly distinguish the drug rash from eczema or inflicted skin lesions.

\vspace{-10pt}
\section{Conclusion}
In this work, we propose the real-world evaluation framework to fairly assess and compare the performance between FSL and CSL-based methods on the all-way skin condition classification problem, where extreme class imbalance exists.
We find that FSL methods do not outperform class imbalance techniques, yet when ensembled with CSL-based class imbalance techniques, they lead to improved model performance especially for the rare classes.
We also observe that ensembling the models with different strengths and more heterogeneity, such as upsampling and FSL, yields promising results.

To further improve the FSL methods, researchers have, in recent times, proposed stronger representation learning strategies via feature reuse~\citep{raghu2019rapid}, or self-supervised learning~\citep{tian2020rethinking}. We believe that evaluating the FSL methods on real-world benchmarks and use cases as the one outlined in this work can greatly accelerate progress in the development of FSL methods with real-world utility.

For future deployment and real-world impact, we also need to account for potential model failures.
For example, false positive cases may lead to wastage of medical resources while false negatives may lead to delayed treatment, especially for malignant cases.
The ensemble of both CSL and FSL approaches may mitigate these issues, yet to assess the generalizability of the results one needs to evaluate our proposed methods on other datasets and domains, which is an interesting future direction.

%\newpage
\acks{The authors thank Dr. Jaehoon Lee for detailed comments and reviews, Dr. Greg Corrado and all members of the Google Health Dermatology team for their support.}

\bibliography{main}

\begin{thebibliography}{29}
\providecommand{\natexlab}[1]{#1}
\providecommand{\url}[1]{\texttt{#1}}
\expandafter\ifx\csname urlstyle\endcsname\relax
  \providecommand{\doi}[1]{doi: #1}\else
  \providecommand{\doi}{doi: \begingroup \urlstyle{rm}\Url}\fi

\bibitem[Breiman(2001)]{breiman2001random}
Leo Breiman.
\newblock Random forests.
\newblock \emph{Machine learning}, 45\penalty0 (1):\penalty0 5--32, 2001.

\bibitem[Chawla et~al.(2002)Chawla, Bowyer, Hall, and
  Kegelmeyer]{chawla2002smote}
Nitesh~V Chawla, Kevin~W Bowyer, Lawrence~O Hall, and W~Philip Kegelmeyer.
\newblock Smote: synthetic minority over-sampling technique.
\newblock \emph{Journal of artificial intelligence research}, 16:\penalty0
  321--357, 2002.

\bibitem[Chen et~al.(2019)Chen, Liu, Kira, Wang, and Huang]{chen2019closer}
Wei-Yu Chen, Yen-Cheng Liu, Zsolt Kira, Yu-Chiang~Frank Wang, and Jia-Bin
  Huang.
\newblock A closer look at few-shot classification.
\newblock \emph{ICLR}, 2019.

\bibitem[Cortes and Vapnik(1995)]{cortes1995support}
Corinna Cortes and Vladimir Vapnik.
\newblock Support-vector networks.
\newblock \emph{Machine learning}, 20\penalty0 (3):\penalty0 273--297, 1995.

\bibitem[Cruz-Roa et~al.(2013)Cruz-Roa, Ovalle, Madabhushi, and
  Osorio]{cruz2013deep}
Angel~Alfonso Cruz-Roa, John Edison~Arevalo Ovalle, Anant Madabhushi, and Fabio
  Augusto~Gonz{\'a}lez Osorio.
\newblock A deep learning architecture for image representation, visual
  interpretability and automated basal-cell carcinoma cancer detection.
\newblock \emph{MICCAI}, 2013.

\bibitem[Eban et~al.(2017)Eban, Schain, Mackey, Gordon, Rifkin, and
  Elidan]{eban2017scalable}
Elad Eban, Mariano Schain, Alan Mackey, Ariel Gordon, Ryan Rifkin, and Gal
  Elidan.
\newblock Scalable learning of non-decomposable objectives.
\newblock \emph{Artificial Intelligence and Statistics}, pages 832--840, 2017.

\bibitem[Esteva et~al.(2017)Esteva, Kuprel, Novoa, Ko, Swetter, Blau, and
  Thrun]{esteva2017dermatologist}
Andre Esteva, Brett Kuprel, Roberto~A Novoa, Justin Ko, Susan~M Swetter,
  Helen~M Blau, and Sebastian Thrun.
\newblock Dermatologist-level classification of skin cancer with deep neural
  networks.
\newblock \emph{Nature}, 542\penalty0 (7639):\penalty0 115--118, 2017.

\bibitem[Finn et~al.(2017)Finn, Abbeel, and Levine]{finn2017model}
Chelsea Finn, Pieter Abbeel, and Sergey Levine.
\newblock Model-agnostic meta-learning for fast adaptation of deep networks.
\newblock \emph{ICML}, 2017.

\bibitem[Ghorbani et~al.(2020)Ghorbani, Natarajan, Coz, and
  Liu]{ghorbani2020dermgan}
Amirata Ghorbani, Vivek Natarajan, David Coz, and Yuan Liu.
\newblock Derm{GAN}: Synthetic generation of clinical skin images with
  pathology.
\newblock \emph{Proceedings of the Machine Learning for Health NeurIPS
  Workshop}, pages 155--170, 2020.

\bibitem[Guo et~al.(2020)Guo, Codella, Karlinsky, Codella, Smith, Saenko,
  Rosing, and Feris]{guo2020broader}
Yunhui Guo, Noel~C Codella, Leonid Karlinsky, James~V Codella, John~R Smith,
  Kate Saenko, Tajana Rosing, and Rogerio Feris.
\newblock A broader study of cross-domain few-shot learning.
\newblock \emph{ECCV}, 2020.

\bibitem[Koch et~al.(2015)Koch, Zemel, and Salakhutdinov]{koch2015siamese}
Gregory Koch, Richard Zemel, and Ruslan Salakhutdinov.
\newblock Siamese neural networks for one-shot image recognition.
\newblock \emph{ICML deep learning workshop}, 2, 2015.

\bibitem[Lake et~al.(2011)Lake, Salakhutdinov, Gross, and
  Tenenbaum]{lake2011one}
Brenden Lake, Ruslan Salakhutdinov, Jason Gross, and Joshua Tenenbaum.
\newblock One shot learning of simple visual concepts.
\newblock \emph{Proceedings of the annual meeting of the cognitive science
  society}, 33\penalty0 (33), 2011.

\bibitem[Latinne et~al.(2001)Latinne, Saerens, and
  Decaestecker]{latinne2001adjusting}
Patrice Latinne, Marco Saerens, and Christine Decaestecker.
\newblock Adjusting the outputs of a classifier to new a priori probabilities
  may significantly improve classification accuracy: evidence from a
  multi-class problem in remote sensing.
\newblock \emph{ICML}, 1:\penalty0 298--305, 2001.

\bibitem[Le et~al.(2020)Le, Le, Ngo, and Ngo]{le2020transfer}
Duyen~NT Le, Hieu~X Le, Lua~T Ngo, and Hoan~T Ngo.
\newblock Transfer learning with class-weighted and focal loss function for
  automatic skin cancer classification.
\newblock \emph{arXiv preprint arXiv:2009.05977}, 2020.

\bibitem[Li et~al.(2020)Li, Yu, Fu, and Heng]{li2019difficulty}
Xiaomeng Li, Lequan Yu, Chi-Wing Fu, and Pheng-Ann Heng.
\newblock Difficulty-aware meta-learning for rare disease diagnosis.
\newblock \emph{MICCAI}, 2020.

\bibitem[Lin et~al.(2017)Lin, Goyal, Girshick, He, and
  Doll{\'a}r]{lin2017focal}
Tsung-Yi Lin, Priya Goyal, Ross Girshick, Kaiming He, and Piotr Doll{\'a}r.
\newblock Focal loss for dense object detection.
\newblock \emph{ICCV}, pages 2980--2988, 2017.

\bibitem[Liu et~al.(2020)Liu, Jain, Eng, Way, Lee, Bui, Kanada,
  de~Oliveira~Marinho, Gallegos, Gabriele, et~al.]{liu2020deep}
Yuan Liu, Ayush Jain, Clara Eng, David~H Way, Kang Lee, Peggy Bui, Kimberly
  Kanada, Guilherme de~Oliveira~Marinho, Jessica Gallegos, Sara Gabriele,
  et~al.
\newblock A deep learning system for differential diagnosis of skin diseases.
\newblock \emph{Nature Medicine}, pages 1--9, 2020.

\bibitem[Mahajan et~al.(2020)Mahajan, Sharma, and Vig]{mahajan2020meta}
Kushagra Mahajan, Monika Sharma, and Lovekesh Vig.
\newblock Meta-{D}erm{D}iagnosis: Few-shot skin disease identification using
  meta-learning.
\newblock \emph{Proceedings of the IEEE/CVF Conference on Computer Vision and
  Pattern Recognition Workshops}, pages 730--731, 2020.

\bibitem[Prabhu et~al.(2019)Prabhu, Kannan, Ravuri, Chaplain, Sontag, and
  Amatriain]{prabhu2019few}
Viraj Prabhu, Anitha Kannan, Murali Ravuri, Manish Chaplain, David Sontag, and
  Xavier Amatriain.
\newblock Few-shot learning for dermatological disease diagnosis.
\newblock \emph{MLHC}, 2019.

\bibitem[Raghu et~al.(2020)Raghu, Raghu, Bengio, and Vinyals]{raghu2019rapid}
Aniruddh Raghu, Maithra Raghu, Samy Bengio, and Oriol Vinyals.
\newblock Rapid learning or feature reuse? towards understanding the
  effectiveness of {MAML}.
\newblock \emph{ICLR}, 2020.

\bibitem[Ravi and Larochelle(2017)]{ravi2016optimization}
Sachin Ravi and Hugo Larochelle.
\newblock Optimization as a model for few-shot learning.
\newblock \emph{ICLR}, 2017.

\bibitem[Seth et~al.(2017)Seth, Cheldize, Brown, and Freeman]{seth2017global}
Divya Seth, Khatiya Cheldize, Danielle Brown, and Esther~E Freeman.
\newblock Global burden of skin disease: inequities and innovations.
\newblock \emph{Current dermatology reports}, 6\penalty0 (3):\penalty0
  204--210, 2017.

\bibitem[Snell et~al.(2017)Snell, Swersky, and Zemel]{snell2017prototypical}
Jake Snell, Kevin Swersky, and Richard Zemel.
\newblock Prototypical networks for few-shot learning.
\newblock \emph{Advances in neural information processing systems}, pages
  4077--4087, 2017.

\bibitem[Sung et~al.(2018)Sung, Yang, Zhang, Xiang, Torr, and
  Hospedales]{sung2018learning}
Flood Sung, Yongxin Yang, Li~Zhang, Tao Xiang, Philip~HS Torr, and Timothy~M
  Hospedales.
\newblock Learning to compare: Relation network for few-shot learning.
\newblock \emph{CVPR}, pages 1199--1208, 2018.

\bibitem[Tian et~al.(2020)Tian, Wang, Krishnan, Tenenbaum, and
  Isola]{tian2020rethinking}
Yonglong Tian, Yue Wang, Dilip Krishnan, Joshua~B Tenenbaum, and Phillip Isola.
\newblock Rethinking few-shot image classification: a good embedding is all you
  need?
\newblock \emph{ECCV}, 2020.

\bibitem[Triantafillou et~al.(2020)Triantafillou, Zhu, Dumoulin, Lamblin, Evci,
  Xu, Goroshin, Gelada, Swersky, Manzagol, et~al.]{triantafillou2019meta}
Eleni Triantafillou, Tyler Zhu, Vincent Dumoulin, Pascal Lamblin, Utku Evci,
  Kelvin Xu, Ross Goroshin, Carles Gelada, Kevin Swersky, Pierre-Antoine
  Manzagol, et~al.
\newblock Meta-{D}ataset: A dataset of datasets for learning to learn from few
  examples.
\newblock \emph{ICLR}, 2020.

\bibitem[Vinyals et~al.(2016)Vinyals, Blundell, Lillicrap, Wierstra,
  et~al.]{vinyals2016matching}
Oriol Vinyals, Charles Blundell, Timothy Lillicrap, Daan Wierstra, et~al.
\newblock Matching networks for one shot learning.
\newblock \emph{Advances in neural information processing systems}, pages
  3630--3638, 2016.

\bibitem[Weng et~al.(2019)Weng, Cai, Lin, Tan, and Chen]{weng2019multimodal}
Wei-Hung Weng, Yuannan Cai, Angela Lin, Fraser Tan, and Po-Hsuan~Cameron Chen.
\newblock Multimodal multitask representation learning for pathology biobank
  metadata prediction.
\newblock \emph{Machine Learning for Health NeurIPS Workshop}, 2019.

\bibitem[Yang et~al.(2018)Yang, Sun, Liang, and Rosin]{yang2018clinical}
Jufeng Yang, Xiaoxiao Sun, Jie Liang, and Paul~L Rosin.
\newblock Clinical skin lesion diagnosis using representations inspired by
  dermatologist criteria.
\newblock \emph{CVPR}, pages 1258--1266, 2018.

\end{thebibliography}
\bibliographystyle{plain}

\clearpage

\appendix

\setcounter{figure}{0} \renewcommand{\thefigure}{A\arabic{figure}}
\setcounter{table}{0} \renewcommand{\thetable}{A\arabic{table}}

\section{Overview of the Class Imbalance Methods}
\subsection{CSL-based class imbalance techniques}
The following list describes the details of class imbalance techniques adopted for the CSL.
\begin{itemize}
\item Upsampling: on top of the CSL baseline with the cross entropy loss, the uniform sampling is performed across all classes during training.
\item Bias initialization (BI)~\citep{lin2017focal}: the final layer of the network, i.e., classification heads, is initialized with bias of the $\log$ values of the amount of training examples in the class.
\item Inverse frequency weighting (IFW): the cross entropy loss is further weighted by the inverse frequency of the amount of training examples in the class.
\item Focal loss (FL)~\citep{lin2017focal}: the cross entropy loss is extended to the $\alpha$-balanced cross entropy loss that accounts for the class imbalance by multiplying a weighting factor $\alpha \in [0, 1]$: $\mathcal{L}_{\alpha\mathrm{CE}} = -\sum_i \alpha_i \log(p_i)$, where $a_i = a$ for correct prediction, and $a_i=1-a$ otherwise. It also attempts to down-weigh easy samples and focus on hard samples by introducing a modulating factor $(1-p_i)^{\gamma}$ with a focusing parameter $\gamma \geq 0$. The objective can be expressed as $\mathcal{L}_\mathrm{focal} = - \sum_i \alpha_i (1-p_i)^{\gamma} \log(p_i)$ , where $M$ is a multiplier.
\item Prior correction: predictions from a CSL baseline model are further divided by maximum-likelihood estimates for class priors based on the training set. This method can be seen as a specific use case of the post-hoc correction method introduced by \citep{latinne2001adjusting} to account for label shift and calibrate the model predictions to a domain with uniform class distribution.

% This method has two interpretations under the assumption of model calibration. First, it can be seen as conversion of inferences $p(y|x)$ to likelihood ratios $p(x|y) / p(x)$ for class $y$ and model inputs $x$. A more helpful interpretation as label-shift correction into a target domain with uniform class distribution.
\end{itemize}
We further combine BI/IFW, as well as FL/IFW, for the combined class imbalance techniques.

\subsection{FSL algorithms}
The methods for FSL can be categorized into batch training and episodic training methods.
For batch learning, we explore the following:
\begin{itemize}
\item k-nearest neighbor ($k$-NN): the $k$-NN method classify the given test example based on the cosine distance between the test example and the class centroids in the vector space.
\item Baseline++~\citep{chen2019closer}: Baseline++ is a finetune model that uses the support set in testing episodes to train the final layer on top of the embeddings.
\end{itemize}
For the episodic training, we select the following meta-learners from the metric-based and optimization-based algorithms:
\begin{itemize}
\item Metric-based algorithms
\begin{itemize}
\item Matching networks~\citep{vinyals2016matching}: matching networks use the average of class example embeddings, which is computed by bilateral long short term memory (LSTM), as class prototypes, and classify query examples based on the cosine distance-weighted linear combination of the support labels.
\item Prototypical networks~\citep{snell2017prototypical}: prototypical networks use the average of embeddings learned from CNN as class prototypes, and classify query examples by computing the Euclidean distance between the embeddings of prototypes and queries.
\item Relation networks~\citep{sung2018learning}: relation networks also average the embeddings to create prototypes, yet concatenate the prototype of each class with query examples, and use the relation module, which is parameterized by additional layers, to predict the class probability given the query.
\end{itemize}
\item Optimization-based algorithms
\begin{itemize}
\item Model Agnostic Meta-Learning (MAML)~\citep{finn2017model}: MAML is an optimization-based method that attempt to learn a set of optimal initialization parameters to fast learn on different downstream tasks with small number of gradient steps. We use the first-order approximation MAML in our study.
\item Proto-MAML~\citep{triantafillou2019meta}: Proto-MAML combines the ideas of prototypical networks and MAML by reintepreting the prototypical network as a linear layer on top of the learned embedding~\citep{snell2017prototypical}. Then the linear layer with prototpical network-equivalent weights and bias can be used as the initialization of the last layer of MAML during training.
\end{itemize}
\end{itemize}

\onecolumn

\begin{table*}[htbp]
\centering
\begin{tabular}{@{}ccccc@{}}
\toprule
Split & Train & Validation & Test &  \\ \midrule
Patient number & 9249 & 302 & 2755 &  \\
Image number & 11403 & 526 & 3136 &  \\ \bottomrule
\end{tabular}
\caption{Statistics of the skin dataset in the study.}
\label{tab-a1}
\end{table*}

\begin{table*}[htbp]
\centering
\resizebox{\linewidth}{!}
{
\small
\begin{tabular}{@{}ccl@{}}
\toprule
 & \#Class & Included classes \\ \midrule
Common & 27 & \begin{tabular}[c]{@{}l@{}}Acne, Actinic Keratosis, Allergic Contact Dermatitis, Alopecia Areata, \\ Androgenetic Alopecia, Basal Cell Carcinoma, Cyst, Eczema, Folliculitis, \\ Hidradenitis, Lentigo, Melanocytic Nevus, Melanoma, \\ Other, Post Inflammatory Hyperpigmentation, Psoriasis, \\ Squamous cell carcinoma/squamous cell carcinoma in situ (SCC/SCCIS), \\ seborrheic keratosis/irritated seborrheic keratosis (SK/ISK), \\ Scar Condition, Seborrheic Dermatitis, Skin Tag, Stasis Dermatitis, \\ Tinea, Tinea Versicolor, Urticaria, Verruca vulgaris, Vitiligo\end{tabular} \\ \midrule
Rare & 42 & \begin{tabular}[c]{@{}l@{}}Abscess, Acanthosis nigricans, Acne keloidalis, Amyloidosis of skin, \\ Central centrifugal cicatricial alopecia, Condyloma acuminatum, \\ Confluent and reticulate papillomatosis, Cutaneous lupus, \\ Dermatofibroma, Dissecting cellulitis of scalp, Drug Rash, \\ Erythema nodosum, Folliculitis decalvans, Granuloma annulare, \\ Hemangioma, Herpes Simplex, Herpes Zoster, \\ Idiopathic guttate hypomelanosis, Inflicted skin lesions, Insect Bite, \\ Intertrigo, Irritant Contact Dermatitis, Keratosis pilaris, \\ Lichen Simplex Chronicus (LSC), Lichen planus/lichenoid eruption, \\ Lichen sclerosus, Lipoma, Melasma, Milia, Molluscum Contagiosum, \\ Onychomycosis, Paronychia, Perioral Dermatitis, Photodermatitis, \\ Pigmented purpuric eruption, Pityriasis rosea, Prurigo nodularis, \\ Pyogenic granuloma, Rosacea, Scabies, Telogen effluvium, Xerosis\end{tabular} \\ \bottomrule
\end{tabular}
}
\caption{Detailed categorization for common and rare skin conditions in the study.}
\label{tab-a2}
\normalsize
\end{table*}

\begin{table*}[htbp]
\centering
\resizebox{\linewidth}{!}
{
\begin{tabular}{@{}cccccc@{}}
\toprule
 & Train & 5-way-5-shot & 10-way-5-shot & 14-way-5-shot & 14-way-9-shot \\ 
Method & Eval & 5-way-5-shot & 5-way-5-shot & 5-way-5-shot & 5-way-5-shot \\ \midrule
k-NN &  & 0.544$\pm$0.010 & \textbf{0.55$\pm$0.010} & 0.548$\pm$0.010 & 0.545$\pm$0.010 \\
Baseline++ &  & 0.475$\pm$0.010 & 0.469$\pm$0.010 & \textbf{0.593$\pm$0.010} & 0.488$\pm$0.010 \\
Matching networks &  & \textbf{0.597$\pm$0.010} & 0.592$\pm$0.010 & 0.566$\pm$0.010 & 0.581$\pm$0.010 \\
Prototypical networks &  & 0.587$\pm$0.010 & \textbf{0.613$\pm$0.010} & 0.610$\pm$0.010 & 0.605$\pm$0.010 \\
Relation networks &  & 0.585$\pm$0.010 & \textbf{0.586$\pm$0.010} & 0.578$\pm$0.010 & 0.559$\pm$0.010 \\
MAML &  & 0.585$\pm$0.010 & \textbf{0.594$\pm$0.010} & 0.587$\pm$0.010 & 0.580$\pm$0.010 \\
Proto-MAML &  & 0.599$\pm$0.010 & \textbf{0.628$\pm$0.010} & 0.626$\pm$0.010 & 0.595$\pm$0.010 \\ \bottomrule
\end{tabular}
}
\caption{Standard FSL evaluation for the skin dataset with the change of $N$ and $k$ value (5-way-5-shot accuracy, with 95\% confidence interval).``14-way-9-shot'' indicates that we use 14-way-9-shot for the training in meta-learning, but use 5-way-5-shot for all evaluations. The boldface values indicate the best setting for each FSL algorithm. We found that higher way is in general helpful to improve the performance, yet the 10-way-5-shot training is relatively optimal for standard FSL using the teledermatology dataset to prevent overfit/underfit while performing meta-learning. Proto-MAML and prototypical networks outperforms other methods.}
\label{tab-a3}
\end{table*}

\begin{table*}[htbp]
\centering
\resizebox{\linewidth}{!}
{
\begin{tabular}{@{}ccccc@{}}
\toprule
Method & 5-way-5-shot & 10-way-5-shot & 14-way-5-shot & 14-way-9-shot \\ \midrule
k-NN & 0.137 (0.094/0.168) & 0.134 (0.098/0.159) & 0.117 (0.101/0.125) & 0.136 (0.093/0.161) \\
Baseline++ & 0.048 (0.044/0.052) & 0.050 (0.044/0.055) & 0.042 (0.041/0.044) & 0.049 (0.039/0.056) \\
Matching networks & \textbf{0.235 (0.097/0.325)} & \textbf{0.228 (0.135/0.302)} & \textbf{0.229 (0.118/0.304)} & \textbf{0.264 (0.130/0.350)} \\
Prototypical networks & 0.211 (0.159/0.244) & 0.219 (0.157/0.262) & 0.219 (0.159/0.261) & 0.213 (0.168/0.245) \\
Relation networks & 0.038 (0.026/0.046) & 0.056 (0.036/0.071) & 0.090 (0.047/0.119) & 0.072 (0.045/0.091) \\
MAML & 0.078 (0.042/0.101) & 0.181 (0.056/0.250) & 0.157 (0.065/0.218) & 0.210 (0.080/0.296) \\
Proto-MAML & 0.123 (0.089/0.146) & 0.184 (0.107/0.237) & 0.213 (0.122/0.274) & 0.240 (0.134/0.308) \\ \bottomrule
\end{tabular}
}
\caption{Real-world FSL evaluation. We report the performance of models trained by different $N$-way-$k$-shot setting. The value outside the parentheses is the balanced accuracy of \emph{ALL} classes, and the values inside the parenthesis are the balanced accuracy of common/rare classes, respectively. The boldface values indicate the best algorithm for each training setup. Different from the standard FSL evaluation, higher $N$ and $k$ yield better meta-learner in the real-world FSL evaluation. The performance of classifying rare classes is consistently better than classifying the common classes. Matching networks and Proto-MAML using 14-way-9-shot training outperforms the other methods under the real-world evaluation. We use them for the method comparison and model ensemble.}
\label{tab-a4}
\end{table*}

\begin{table*}[htbp]
\centering
\begin{tabular}{@{}ccccc@{}}
\toprule
 & All-way accuracy & \begin{tabular}[c]{@{}c@{}}Balanced\\ accuracy\end{tabular} & \begin{tabular}[c]{@{}c@{}}Balanced acc.\\ (common class)\end{tabular} & \begin{tabular}[c]{@{}c@{}}Balanced. acc.\\ (rare class)\end{tabular} \\ \midrule
FSL & 0.147 & 0.264 & 0.130 & 0.359 \\
CSL baseline & 0.648 & 0.411 & 0.525 & 0.347 \\
Upsampling & 0.610 & 0.504 & \textbf{0.575} & 0.458 \\
BI & \textbf{0.661} & 0.420 & 0.537 & 0.345 \\
IFW & 0.621 & 0.494 & 0.523 & 0.476 \\
BI/IFW & 0.620 & 0.506 & 0.515 & 0.500 \\
FL & 0.657 & 0.429 & 0.530 & 0.364 \\
FL/IFW & 0.395 & 0.552 & 0.455 & 0.616 \\
Prior Correction & 0.450 & \textbf{0.580} & 0.518 & \textbf{0.620} \\ \bottomrule
\end{tabular}
\caption{Comparison between FSL, CSL baseline and other class imbalance techniques in the all-way classification problem under single model (non-ensemble) setting. The boldface values indicate the best method for each evaluation metric. Note that all-way accuracy is a biased metric towards common classes, as the test set is dominated with common classes. Also note that we include here an additional class-imbalance method "Prior Correction" (for details please refer to Appendix A.1). Prior correction achieves the greatest balanced accuracy, an observation theoretically supported by it's interpretation as a label-shift correction and the fact that balanced accuracy is equivalent to overall accuracy in a domain with uniform class distribution. Single FSL model (matching networks) is not beneficial while using it independently for all-way classification. Instead, single CSL model with upsampling strategy outperforms others in the common class classification, and single CSL model with focal loss/inverse frequency weighting yields the best performance for the rare class classification.}
\label{tab-a5}
\end{table*}

\begin{table*}[htbp]
\centering
\resizebox{\linewidth}{!}
{
\begin{tabular}{@{}cccccc@{}}
\toprule
\#Model & Combination & \begin{tabular}[c]{@{}c@{}}All-way\\ accuracy\end{tabular} & \begin{tabular}[c]{@{}c@{}}Balanced\\ accuracy\end{tabular} & \begin{tabular}[c]{@{}c@{}}Balanced acc.\\ (common class)\end{tabular} & \begin{tabular}[c]{@{}c@{}}Balanced acc.\\ (rare class)\end{tabular} \\ \midrule
2 & 2 FSL & 0.233 & 0.340 & 0.215 & 0.421 \\
 & 2 CSL Baseline & 0.663 & 0.431 & 0.523 & 0.371 \\
 & FSL+CSL Baseline & 0.633 & 0.435 & 0.506 & 0.389 \\
 & 2 Upsampling & 0.617 & 0.509 & 0.58 & 0.464 \\
 & FSL+Upsampling & 0.582 & 0.524 & 0.557 & 0.503 \\ \midrule
4 & 4 FSL & 0.24 & 0.363 & 0.216 & 0.459 \\
 & 4 Upsampling & 0.623 & 0.529 & 0.591 & 0.489 \\
 & FSL+Upsampling+BI/IFW+FL/IFW & 0.582 & 0.558 & 0.541 & 0.569 \\ \bottomrule
\end{tabular}
}
\caption{Comparison between ensemble FSL, CSL baseline and CSL-based class imbalance techniques in the all-way classification problem under ensemble setting. Note that all-way accuracy is a biased metric towards common classes, as the test set is dominated with common classes. In all-way classification, the balanced accuracy on all classes improves after ensembling with the FSL model (2 CSL baseline versus FSL+CSL baseline, and 2 upsampling versus FSL+upsampling). Similarly, combining models with different learning mechanisms helps improve the balanced accuracy across all classes (4 FSL versus 4 upsampling versus 4 different models).}
\label{tab-a6}
\end{table*}

\end{document}